\documentclass[conference]{IEEEtran}
\pdfoutput=1
\IEEEoverridecommandlockouts
\usepackage{cite}
\usepackage{amsmath,amssymb,amsfonts}
\usepackage{algorithmic}
\usepackage{graphicx}
\usepackage{textcomp}
\usepackage{xcolor}
\usepackage{mathtools}
\usepackage{multirow}
\usepackage{caption}
\usepackage{booktabs}
\usepackage{hyperref}
\usepackage{amsfonts,amssymb}
\usepackage{algorithm}
\usepackage{algorithmic}
\usepackage{fancyhdr}

\def\BibTeX{{\rm B\kern-.05em{\sc i\kern-.025em b}\kern-.08em
    T\kern-.1667em\lower.7ex\hbox{E}\kern-.125emX}}
\begin{document}

\title{EC-LDA : Label Distribution Inference Attack against Federated Graph Learning with Embedding Compression\\
% {\footnotesize \textsuperscript{*}Note: Sub-titles are not captured in Xplore and
% should not be used}
% \thanks{Identify applicable funding agency here. If none, delete this.}
}

\author{
\IEEEauthorblockN{
Tong Cheng\IEEEauthorrefmark{2},
Jie Fu\IEEEauthorrefmark{3}\IEEEauthorrefmark{1},
% \thanks{$^{*}$Chuan Qin and Haiping Ma are corresponding authors.},
Xinpeng Ling\IEEEauthorrefmark{2},
Huifa Li\IEEEauthorrefmark{2},
Zhili Chen\IEEEauthorrefmark{2}\IEEEauthorrefmark{1}\thanks{$^{}$* Jie Fu and Zhili Chen are co-corresponding authors.},
Haifeng Qian\IEEEauthorrefmark{2},
Junqing Gong\IEEEauthorrefmark{2}
}
\IEEEauthorblockA{
\IEEEauthorrefmark{2}Software Engineering Institute, East China Normal University, Shanghai, China}

\IEEEauthorblockA{
\IEEEauthorrefmark{3}Department of Computer Science, Stevens Institute of Technology, New Jersey, USA\\
\{tcheng, xpling, huifali\}@stu.ecnu.edu.cn, jfu13@stevens.edu, \{zhlchen, jqgong\}@sei.ecnu.edu.cn, hfqian@admin.ecnu.edu.cn}
}

	% \affiliation{%
	% 	\institution{$^1$Shanghai Key Laboratory of Trustworthy Computing, East China Noraml University, China}
	% }
	% \affiliation{%
	% 	\institution{$^2$Department of Electronic and Information Engineering, The Hong Kong Polytechnic University, China}
	% }
	% \email{jie.fu@stu.ecnu.edu.cn, qqing.ye@polyu.edu.hk, haibo.hu@polyu.edu.hk, zhlchen@sei.ecnu.edu.cn}
 %    \email{luluwang@stu.ecnu.edu.cn, 10204804424@stu.ecnu.edu.cn, qi-xun.ran@connect.polyu.hk}

\maketitle

\thispagestyle{fancy}
\fancyhead{}
\chead{\textcolor{red}{\large This paper has been accepted by 2025 IEEE International Conference on Data Mining (ICDM 2025)}}

\begin{abstract}
Graph Neural Networks (GNNs) have been widely used for graph analysis. Federated Graph Learning (FGL) is an emerging learning framework to collaboratively train graph data from various clients. Although FGL allows client data to remain localized, a malicious server can still steal client private data information through uploaded gradient. In this paper, we for the first time propose {\em label distribution attacks} (LDAs\footnote{The term ``LDA" here is different from other machine learning terms like Latent Dirichlet Allocation.}) on FGL that aim to infer the label distributions of the client-side data. Firstly, we observe that the effectiveness of LDA is closely related to the variance of node embeddings in GNNs. Next, we analyze the relation between them and propose a new attack named EC-LDA, which significantly improves the attack effectiveness by compressing node embeddings. Then, extensive experiments on node classification and link prediction tasks across six widely used graph datasets show that EC-LDA outperforms the SOTA LDAs. Specifically, EC-LDA can achieve the Cos-sim as high as 1.0 under almost all cases. Finally, we explore the robustness of EC-LDA under differential privacy protection and discuss the potential effective defense methods to EC-LDA. Our code is available at https://github.com/cheng-t/EC-LDA.
% \href{https://anonymous.4open.science/r/EC-LDA}{}.
\end{abstract}


\section{Introduction}
\subsection{Background}
Graph Neural Networks (GNNs), designed to process graph-structured data, have gained significant attention for their effectiveness across various applications including recommendation systems \cite{wang2024distributionally}, social networks \cite{han2024topology}, and protein interaction prediction \cite{jha2022prediction}. GNNs can capture information between neighboring nodes, enhancing the expressiveness of node embeddings and making them highly suitable for real-world applications.
% Graph Neural Networks (GNNs) are designed to process data structured as graphs, distinguishing them from traditional deep learning models that typically handle regular data types such as images and text. GNNs have gained significant attention due to their effectiveness in various applications, including recommendation systems \cite{he2020lightgcn}, social networks \cite{fan2019graph}, and protein interaction prediction \cite{jha2022prediction}. GNNs can capture information between neighboring nodes, enhancing the expressiveness of node embeddings and making them highly suitable for real-world applications.

The performance of GNNs requires a large amount of data. However, due to privacy concerns and regulatory restrictions, machine learning platforms cannot access raw data directly, which makes centralized GNN learning challenging. In recent years, a lot of works \cite{he2021fedgraphnn,zhang2021federated} have integrated Federated Learning (FL) \cite{mcmahan2017communication} with GNNs, proposing the Federated Graph Learning (FGL). FL is a distributed, privacy-preserving machine learning paradigm that enables clients to train models collaboratively while keeping their local data isolated. It addresses the challenge of data silo, where data is distributed across different sources and cannot be easily combined for joint analysis, by enabling model training without the need to share the raw data. 

\begin{figure}[ht]
    \centering
    \includegraphics[width=0.8\linewidth]{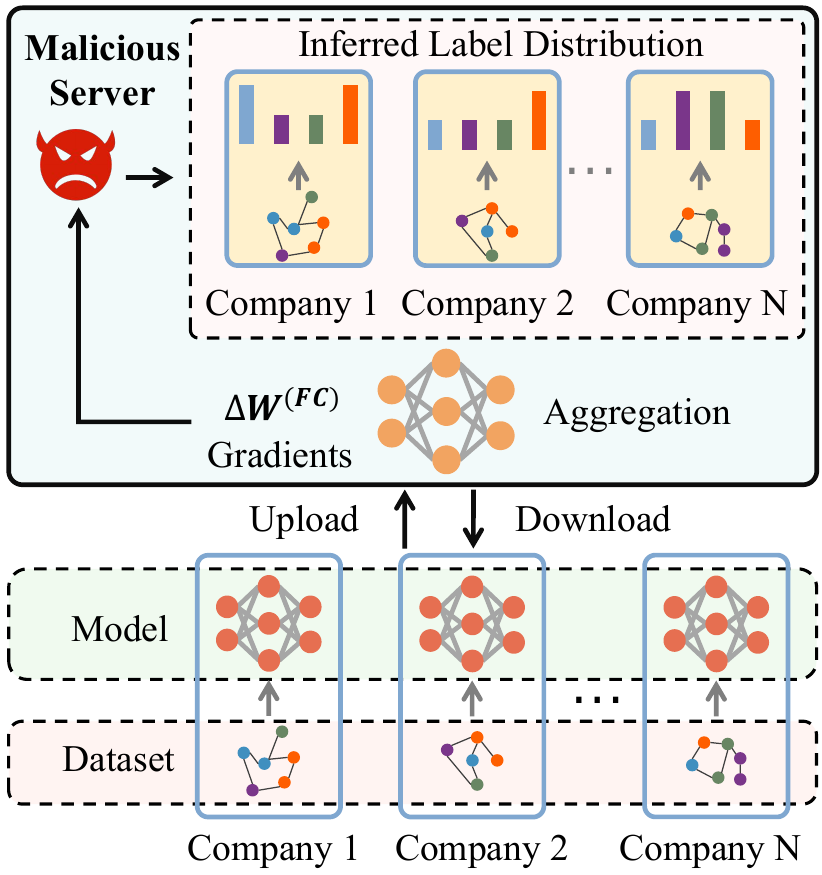}
    \caption{Threat model of EC-LDA.}
    \label{fig_overview}
\end{figure}

However, recent studies have shown that local data in FL remains vulnerable to various attacks \cite{zhu2019deep,geiping2020inverting,fu2024differentially,li2025coward}. One such attack is label distribution inference attacks (LDAs)~\cite{gu2023ldia,wainakh2021user}, which aim to infer the label distribution of client-side local training data by analyzing the gradients shared between clients and the server. This represents a significant privacy threat in FL. 
For example, if multiple online shopping companies collaborate to train a recommendation system model (as shown in Fig. \ref{fig_overview}), a malicious server with access to the label distribution of a private social network could target specific users, increasing the success rate of fraudulent activities and posing a serious threat to user privacy.

\subsection{Previous Works and Limitations}
Despite existing works on LDAs~\cite{gu2023ldia,wainakh2021user,yin2021see,ma2023instance}, they all under non-GNN settings and have many limitations. For example, their attack methods require additional auxiliary datasets~\cite{gu2023ldia,wainakh2021user} or specific activation functions~\cite{yin2021see} as assumption. Additionally, the performance of attack method declines as the local epochs in FL increase, achieving effective results only with a single local training epoch~\cite{ma2023instance,wainakh2021user}. These assumptions restrict their applicability to GNN scenarios. Furthermore, the message-passing characteristic among adjacent nodes in GNNs complicates node embeddings, potentially introducing irrelevant details into the model output, thereby making LDA in GNNs more challenging.

\subsection{Our Contributions} In this paper, we propose Embedding Compression-Label Distribution Inference Attack (EC-LDA), a novel LDA in FGL. Through exploring the factors behind the poor performance of LDAs in GNNs, we find there are strong correlation between the effectiveness of LDA and the variance of node embeddings. In GNNs, the variance in node embeddings primarily originates from the GNN layers and increases with the number of layers, which in turn degrades the attack performance of LDA. EC-LDA addresses this issue by compressing node embeddings, thereby reducing their variance and enhancing attack performance. Moreover, EC-LDA overcomes the drawbacks of existing LDAs. It keeps stable performance as training samples and local epochs increase and doesn't need specific activation functions or extra datasets for attacker. Our main contributions are as follows:
\begin{itemize}
    \item We analyze the relationship between the performance of LDA and the message-passing characteristics of GNNs, and introduce EC-LDA, the first approach that  implement efficient LDA on FGL.
    
    \item We apply EC-LDA to six graph datasets and conduct extensive experiments targeting both node classification and link prediction tasks. These experiments demonstrate that EC-LDA consistently achieves significant attack performance across various scenarios. EC-LDA achieves a Cos-sim as high as 1.00 under almost all cases.
    
    \item We utilize DP-GNN (node-level differential privacy) and Label-DP (label-level differential privacy) for local GNN training and evaluate the robustness of EC-LDA. We further discuss potential effective defense methods for EC-LDA.
\end{itemize}

\subsection{Paper Structure}
% The remainder of this paper is organized as follows: The next section introduces preliminaries on GNNs, FL, and the threat model. Subsequently, we analyze the GNN message passing mechanism in LDA and present our proposed EC-LDA approach. Extensive experiments validate the effectiveness of EC-LDA and discuss defense strategies. We then detail detection methods for EC-LDA. Finally, We present related work and conclusion.
The rest of the paper is organized as follows. Section \ref{sec:2} introduces the preliminaries knowledge and the threat model. In Section \ref{sec:3}, we analyze the message-passing in LDA and give the key observation. Though the findings, we proposed our attack EC-LDA approach in Section~\ref{sec:4}. The experimental results are presented in Section \ref{sec:5}. Section \ref{sec:6} reviews related work, followed the limitations in Section \ref{sec:7}. Lastly, conclusion in Section \ref{sec:8}.

\section{Preliminaries}\label{sec:2}
%In this section, we provide a comprehensive introduction to Graph Neural Networks and Federated Learning, while also introducing the threat model of this paper.

\subsection{Graph Neural Networks} In general, GNNs are designed to process data \(G(V,\mathbb{E})\) structured as a graph.
% to learn representations of the graph. 
Here, \(V\) represents the set of nodes and \(\mathbb{E}\) represents the adjacency matrix of \(G\). Each node \(v_{i} \in V\) has a feature vector \(u_{i}\).
GNNs generate valuable node embeddings via message-passing, making them suitable for various downstream tasks like node classification and link prediction.
In this paper, we consider Graph Convolutional Networks (GCN) \cite{kipf2016semi}, Graph Attention Networks (GAT) \cite{velivckovic2017graph}, and GraphSAGE \cite{hamilton2017inductive} as target GNNs.

GNNs typically follow the message-passing strategy that updates the features of nodes iteratively by aggregating the features of their neighbors. Typically, a GNN model's \(h\)-th layer can be formulated as:
\begin{equation}\label{equ_gnn}
u_{i}^{h}=\sigma(u_{i}^{h-1},AGG(u_{j}^{h-1},j\in{\mathbb{B}}_{i})),
\end{equation}
where \(u_{i}^{h-1}\) is the representation obtained at the \((h\! -\!1)\)-th layer of node \(v_{i}\), and \(u_{i}^{0}\) is the node feature \(u_{i}\) of node \(v_{i}\), \({\mathbb{B}}_{i}\) represents the neighbors of node \(v_i\), \(AGG(\cdot)\) represents the aggregation function, \(\sigma\) represents the activation function such as \(ReLU\).

\begin{table*}[ht]
\caption{\(Err\) with the variance of \(I\) when local epochs \(E\) is set to 1 (WikiCS dataset and a 2-layer GCN model).}
\centering
\begin{tabular}{@{}ccccccccccc@{}}
\toprule
Variance*1000 of \(I\) & 41.052 & 23.719 & 21.283 & 10.640 & 4.547 & 2.031 & 0.817 & 0.224 & 0.067 & 0.003 \\ \midrule
\(Err\)                   & 4.367  & 3.433  & 2.765  & 2.217  & 1.523 & 1.012 & 0.599 & 0.309 & 0.138 & 0.028 \\ \bottomrule
\end{tabular}

\label{tab:var_var}
\end{table*}
\subsection{Federated Graph Learning}

A typical FGL system follows the FedAvg \cite{mcmahan2017communication} algorithm. Specifically, the server sends an initial global model to all the clients. Then each client trains a local model with its local private data $G_{i}(V_{i},\mathbb{E}_{i})$ and shares its local model parameters with the server. The server then aggregates the local model parameters of all the clients to construct the global model's parameters, which can be formulated as:
\begin{equation}\label{equ_FL_agg}
W^{t}=\sum_{i=1}^{N}p_{i}W_{i}^{t},
\end{equation}
where $N$ is the number of clients, \(W_{i}^{t}\) and \(p_{i}\) are trained model parameters at the $t$-th epoch and the weight of client \(i\), respectively, \(W^t\) is the global model parameters at the \(t\)-th epoch. The optimization problem of FGL is formulated as:
\begin{equation}\label{equ_FL_goal}
W^{*}=\underset{W}{\operatorname*{argmin}}\sum_{i=1}^{N}p_{i}\mathcal{L}(W,V_{i},\mathbb{E}_{i};Y_{i}),
\end{equation}
where \(\mathcal{L}(\cdot)\) represents the loss function, \(Y_{i}\) is node label in the node classification task or link label in the link prediction task of client \(i\), respectively, \(V_{i}\) represents the set of nodes and \(\mathbb{E}_{i}\) represents the adjacency matrix of \(G_{i}\).

\subsection{Threat Model}
% As shown in Figure~\ref{fig_overview}, we consider the FGL scenario where the server is malicious. The server is interested in the label distributions of the clients' privacy data and has the ability to adjust the parameters of the deployed model, as well as analyze the gradients uploaded by the clients.
As shown in Fig.~\ref{fig_overview}, we consider the FGL scenario where a malicious server is curious about local private data and attempt to reconstruct the label distributions of the clients private data. For instance, consider multiple hospitals collaboratively training a GNN model to predict patients' disease probabilities. Each hospital maintains local patient data graphs where nodes represent patients, edges correspond to medical relationships, and labels denote diagnosis results. By inferring hospital-specific label distributions, attackers could obtain sensitive regional health statistics, potentially revealing disease prevalence in specific healthcare jurisdictions.

\textbf{Attacker's Knowledge and Capability.}
In this scenario, the malicious server adversary can not only observe all gradient updates uploaded by clients but also actively tamper with global model parameters or the aggregation process~\cite{9109557,10536902,pasquini2022eluding}.

% \textbf{Potential Privacy Threats.} 
% Knowledge of clients' private label distribution information by attackers may pose severe privacy threats. 
% In this scenario, the server is not only interested in accessing the label distributions of the clients' private data, but also has the ability to manipulate the parameters of the deployed model. Furthermore, the server can analyze the gradients uploaded by the clients.

% The label distribution inference attack is a type of label recovery attack. 
% \cite{zhu2019deep} was the first to restore the training sample from gradients. They restored the input data and associated label from the gradients using gradient-matching. This method continuously optimizes the dummy input data and associated label by minimizing the mean square error of the gradients of the dummy sample with respect to the true gradients.
% \cite{zhao2020idlg,yin2021see,geng2021towards}

\section{Message-Passing Issue in LDA}\label{sec:3}
In this section, we will analyze how the message-passing mechanism in GNNs exacerbates LDA. We begin with an introduction to LDA. The prior analysis \cite{geng2021towards} has shown that, for the \(k\)-th sample in a batch: 
\(\Delta W_{l}^{(FC)}/E\approx\frac{1}{K}\sum_{k}(p_{k,l}\sum_{m}I_{k,m}-y_{k,l}\sum_{m}I_{k,m})\),
% \Wendy{The notations need to be explained. What are $W_{l}^{(FC)}$, $p_{k,l}$, and  $I_{k,m}$?}
where \(\Delta W_{l}^{(FC)}\) is the sum of the gradients of the \(l\)-th output unit in the last fully connected layer along the input dimension, \(I_{k,m}\) is the input of the \(k\)-th sample at the \(m\)-th input unit of the last fully connected layer, \(p_{k,l}\) is the post-softmax probability at index \(l\) of the \(k\)-th sample, \(E\) is the local epochs.
Furthermore, we have: 
\(\sum_{k}y_{k,l}\cdot \bar{I}\approx\sum_{k}p_{k,l}I_{k}-K\Delta W_{l}^{(FC)}/E\), where \(I_{k}=\sum_{m}I_{k,m}\), and \(\bar{I}\) is the mean value of \(I_{k}\).
% where \(\Delta W_{l}^{(FC)}\) is the sum of the gradients of the \(l\)-th output units of the last fully connected layer along the input dimension, \(I_{k,m}\) is the input of the \(k\)-th sample at the \(m\)-th input unit of the last fully connected layer, \(p_{k,l}\) is the post-softmax probability at index \(l\) of the \(k\)-th sample, \(E\) is the local epochs.

% \begin{equation}\label{equ_8}
% \begin{aligned}
% \sum_{k}y_{k,l}\cdot \bar{I}\approx\sum_{k}p_{k,l}I_{k}-K\Delta W_{l}^{(FC)}/E,
% \end{aligned}
% \end{equation}
% where \(I_{k}\) is the sum of \(I_{k,m}\) along \(m\) dimension, \(\bar{I}\) is the mean value of \(I_{k}\). 
In order to separate out \(\sum_{k}y_{k,l}\), we assume that \(I_{k}\) is close to its mean value, i.e., \(\bar{I}\). Based on this assumption, we have:
\begin{equation}\label{equ_10}
\begin{aligned}
d_{l}
&:=\sum_{k}y_{k,l}\\
&\approx\frac{\sum_{k}p_{k,l}I_{k}}{\bar{I}}-\frac{K\Delta W_{l}^{(FC)}}{E\bar{I}},
\end{aligned}
\end{equation}
% where \(\Delta W_{n}^{(FC)}\) is the sum of \(\Delta W_{m,n,k}^{(FC)}\) along \(m\) dimension, \(\bar{I_{k}}\) is the mean value of \(\sum_{m}I_{k,m}\). To simplify the calculation, we assume that \(\sum_{m}I_{k,m}\) is close to its mean value, i.e., \(\bar{I_{k}}\). Based on this assumption, we have:
% \begin{equation}\label{equ_7}
% \begin{aligned}
% d_n&:=\sum_ky_{k,n}\\
% &\approx\sum_kp_{k,n}-\frac{K\Delta W_n^{(FC)}}{\bar{I_k}},
% \end{aligned}
% \end{equation}
where \(d_{l}\) is the number of nodes inferred with label \(l(l=1,2,\ldots,L)\). With the number of each label, we can compute the label distribution, which we define as \(D\). 
% To calculate \(d_{l}\), we need to acquire the five components: \(p_{k,l}\), \(I_{k}\), \(\bar{I}\), \(K\), and \(\Delta W_l^{(FC)}\).
In FL, we assume that the server can get \(E\) and \(\Delta W_l^{(FC)}\) from clients. Additionally, the dummy training data randomly generated by the server can be used to estimate \(K\), \(p_{k,l}\), \(I_{k}\), and \(\bar{I}\). We named this method as LDA. 
% LDA only requires that the last layer of the model be fully connected and that the local model be trained with the cross-entropy loss function.

\begin{figure}[htbp]
    \centering
    \includegraphics[width=\linewidth]{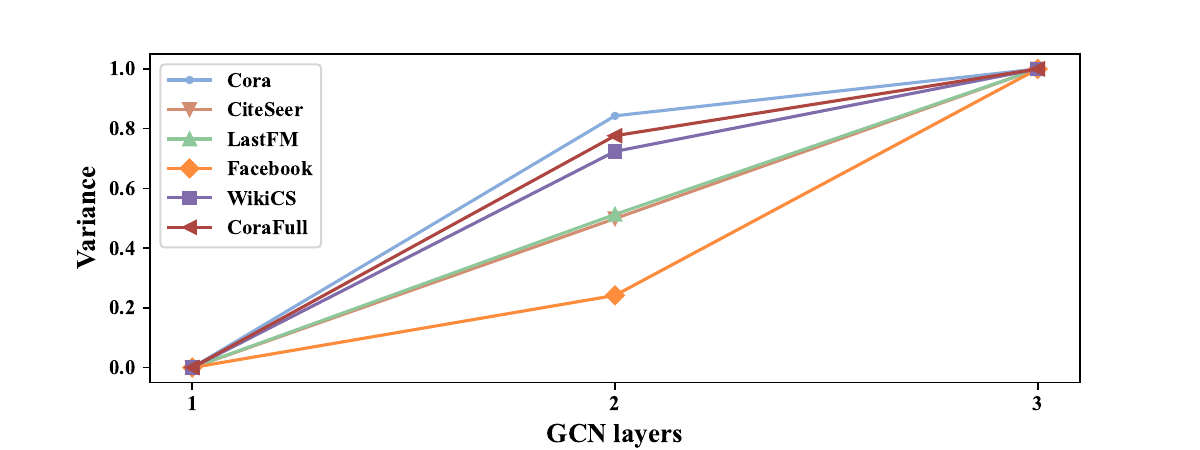}
    \caption{Illustrates the trend of variance of \(I\) with the number of GCN layers. The experimental results at this point are the normalized results.}
    \label{all_dataset_var_metrics_hops}
\end{figure}

% \begin{algorithm}[tb]
% \caption{Label Distribution Recovery Algorithm}
% \label{alg_LDA_gnns}
% \textbf{Input}: Gradient of final fully connected layer \(\Delta W^{(FC)}\), model \(M\), number of classes \(N\), local epochs \(E\)\\
% \textbf{Output}: Label distribution \(D\)
% \begin{algorithmic}[1]
% \STATE Randomly initialize dummy graph \(G^{'}\) with \(K^{'}\) nodes.
% \STATE \((\sum_{k^{\prime}}p_{k^{\prime},1},\ldots,\sum_{k^{\prime}}p_{k^{\prime},N}),I_{k^{\prime}},\bar{I}^{\prime}\leftarrow M(G^{\prime})\)\\
% \FOR {\(i = 1,2,3,\ldots,N\)}
% \STATE Calculate \(d_{i}\) based on Eq. \ref{equ_10}
% \ENDFOR
% \STATE \(S\leftarrow\sum_{n}d_{n}\)\\
% \STATE \(D\leftarrow[d_1/S,d_2/S,d_3/S,\ldots,d_N/S]\) 
% \STATE \textbf{return} \(D\)
% \end{algorithmic}
% \end{algorithm}

\begin{figure*}[ht]
    \centering
    \includegraphics[width=0.7\linewidth]{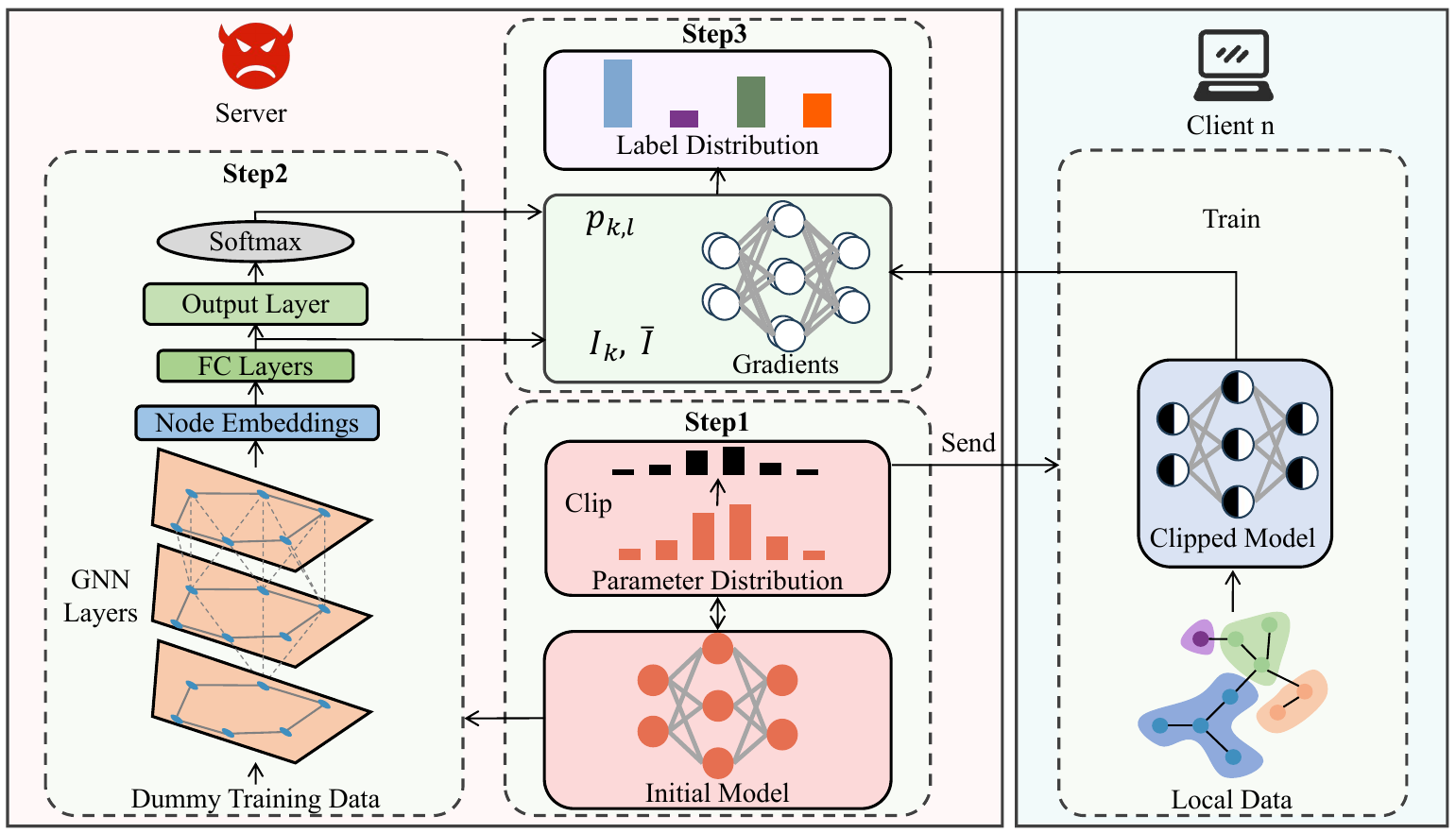}
    \caption{Illustration of the EC-LDA.}
    \label{fig_dataflow}
\end{figure*}

% \begin{figure}[ht]
%     \centering
%     \includegraphics[width=\linewidth]{figure/var_hops.pdf}
%     \caption{Illustrates the trend of variance of \(I\) with the number of GCN layers. The experimental results at this point are the normalized results.}
%     \label{all_dataset_var_metrics_hops}
% \end{figure}

% {\bf Experiments.} To test if LDA can recover the label distribution of training data from gradients, we train 1-3 GCN layers \Wendy{Do you mean 3 GCN models, each of 1, 2, and 3 layers, respectively? Or do you mean 1 GCN of 3 layers?} followed by a fully connected layer, using the SGD optimizer. We perform the attack in the middle round of training. The attack performance for each round is taken as the average of all the clients' attack performance. The final attack performance is averaged over five experiments with different random seeds.

% 这一段要不要再说
% Table X shows the effect of the attack,which we can see that the performance of the attack keeps getting worse as the number of GCN layers increases and the attack is not very effective.

Now we apply and analyze the LDA in GNNs. For the sake of description, we define \(I:=[I_{1},I_{2},\cdot,I_{K}]\). Starting from the above assumption, i.e., \(I_{k}\) is close to \(\bar{I}\), we can find that the error arises mainly because we replace \(I_{k}\) with \(\bar{I}\), i.e., we consider that \(\sum_{k}y_{k,l}I_{k}\approx\sum_{k}y_{k,l}\cdot\bar{I}\). We define the magnitude of the error as:
% \begin{equation}\label{err_0}
% \begin{aligned}
% Err:=\frac{1}{K}\sum_{n}\lvert\sum_{k}y_{k,n}^{*}I_{k}-\sum_{k}y_{k,n}\cdot\bar{I}\rvert.
% \end{aligned}
% \end{equation}
% We assume that \(y_{k,n}^{*}=y_{k,n}\),  in that way:
\begin{equation}\label{err_1}
\begin{aligned}
Err
&:=\frac{1}{K}\sum_{l}\lvert\sum_{k}y_{k,l}(I_{k}-\bar{I})\rvert \\
&=\frac{1}{K}\sum_{k}\lvert I_{k}-\bar{I}\rvert.
\end{aligned}
\end{equation}

We can intuitively see that if the variance of \(I\) is smaller, the \(Err\) is smaller and the performance of the attack is better, and vice versa. To verify our conjecture, we conduct experiments on FGL by performing attacks during mid-training round and analyze the variance of \(I\) versus \(Err\).
Table \ref{tab:var_var} illustrates how \(Err\) varies with the variance of \(I\), demonstrating that the \(Err\) declines as \(I\)'s variance declines, which confirms our suspicions. 

As mentioned earlier, GNNs incorporate neighbor information through the message-passing mechanism. However, the message-passing mechanism also increases the variance of node embeddings. Moreover, the variance of the node embeddings increases with the number of GNN layers in the global model, and the node embeddings are directly related to \(I\). We conduct experiments with GNN models featuring 1, 2, and 3 GCN layers and observe how the variance of \(I\) changes with the number of GCN layers. Fig. \ref{all_dataset_var_metrics_hops} illustrates the variance of \(I\) as a function of the number of GCN layers.
We scaled the variance to [0,1] through min-max normalization: 

\begin{equation}\label{normalization}
\begin{aligned}
V_{new} = \frac{V-V_{min}}{V_{max}-V_{min}},
\end{aligned}
\end{equation}

which $V$ is the variance of \(I\). 
This normalization does not change the trend that the variance in all datasets increases with the number of GNN layers.
% This normalization preserves the relative trends of variance variation across GNN layers in all datasets.
The empirical results indicate a positive correlation between the depth of GCN layers and the variance of \(I\). Consequently, optimizing attack efficacy necessitates minimizing the variance of \(I\).
% It can be seen that as the number of GCN layers increases, the variance of \(I\) becomes larger. Therefore, minimizing the variance of \(I\)  is crucial for enhancing attack efficacy. 

% Note that \(K\) in Eq. \ref{equ_10} is the number of nodes of the dummy training data but not the number of nodes of the clients.

% When \(E\) is greater than 1, we have:
% \begin{equation}\label{equ_10}
% \begin{aligned}
% d_{l}\approx\frac{\sum_{k}p_{k,l}I_{k}}{\bar{I}}-\frac{K\Delta W_{l}^{(FC)}}{E\bar{I}}.
% \end{aligned}
% \end{equation}

% The experiment proves that even with dummy gradients \(\Delta W_{n}^{(FC)}/E\), we can still effectively obtain the label distribution information.

% With Eq. \ref{equ_10}, we can get the number of nodes labeled \(l(l=1,2,\ldots,L)\), \(S:=\sum_{l=1}^Ld_l\) is defined as the sum of all \(d_{l}(l=1,2,\ldots,L)\). Finally, we obtain the inferred labeling distribution \(D:=[d_1/S,d_2/S,\ldots,d_L/S]\). We named this method as LDA.
% (\textbf{L}abel \textbf{D}istribution Inference \textbf{A}ttack). 

% The algorithm flow is illustrated in Algorithm \ref{alg_LDA_gnns}.
\begin{algorithm}[tb]
\caption{Compression of node embeddings}
\label{clip_parameters}
\textbf{Input}: Global model parameters \(G\), clipping threshold \(C\) \\
\textbf{Output}: The global model parameters \(G'\) after clipping and compression
% \Wendy{Maybe "after clipping and compression"?}

\begin{algorithmic}[1]
% \STATE Initialize \(N\) to 0\\
\STATE \(\mathbb{N} \leftarrow 0, G'\leftarrow G \)\\
\FOR{\(p \in G\)}
\STATE \(\mathbb{N}=\mathbb{N}+\Vert p \Vert_2^{2}\)
\ENDFOR
\STATE $\mathbb{N}\leftarrow \mathbb{N}^{1/2}$\\
\FOR{\(p \in G, p' \in G'\)}
\STATE $p' \leftarrow p /max(1,\frac{\mathbb{N}}{C})$\\
\ENDFOR
\STATE \textbf{return} \(G'\)
\end{algorithmic}
\end{algorithm}

\begin{algorithm}[ht]
\caption{EC-LDA against FGL (Server-side)}
\label{active_attack_algo}
\textbf{Input}: Client number \(N\), global rounds \(R\), attack rounds \(A\)\\
\textbf{Output}: Final global model \(W_{R}\), attack result \(\mathbb{L}\)

\begin{algorithmic}[1]
\STATE Initialize global model $W_{0}$, attack result $\mathbb{L}$ \\
\FOR{\(r = 1,2,3,\ldots,R\)}
\IF{\(r \in A\)}
\STATE \colorbox[RGB]{255,229,229}{$W_{r-1}^{'} \leftarrow W_{r-1}$, clipping $W_{r-1}$ with Algorithm \ref{clip_parameters}}  \\
\STATE \colorbox[RGB]{255,229,229}{Generate dummy training data $D_{dummy}$}
\STATE \colorbox[RGB]{255,229,229}{Input $D_{dummy}$ into $W_{r-1}^{'}$ to obtain $p_{k,l}$, $I_k$ and $\bar{I}$}
\ENDIF
\FOR{\(n = 1,2,3,\ldots,N\)}
% \STATE $\Delta W_{r,n} \leftarrow$ \textbf{ClientUpdate($W_{r-1}$)}
% \STATE $ W_{r,n} \leftarrow$ \textbf{ClientUpdate($W_{r-1}$)}, client \(n\) upload the gradients
\STATE Client \(n\) performs local training and uploads $ W_{r,n}$ and gradients
\IF{\(r\) $\in$ \(A\)}
% \STATE Calculate $D$ based on Algorithm \ref{alg_LDA_gnns} \\
\STATE \colorbox[RGB]{255,229,229}{Calculate $D$ based on Equation \ref{equ_10}} \\
\STATE \colorbox[RGB]{255,229,229}{Add $D$ to $\mathbb{L}$}\\
\ENDIF
\ENDFOR
\STATE Server aggregates all local models $W_{r,n}$ to $W_{r}$ \\
\IF{\(r \in A\)}
\STATE \colorbox[RGB]{255,229,229}{Replace $W_{r}$ with the initial model $W_{r-1}^{'}$} \\
\ENDIF
\ENDFOR

% \STATE \textbf{ClientUpdate}:\\
% \STATE Client has local dataset \(G\) and local epoch \(E\)  \\
% \STATE Receive current global model \(M\) from server \\
% \STATE Train the model with the local dataset \(G\) for \(E\) epochs \\
% \STATE Obtain the trained model \(W\) \\
% % \STATE Obtain the gradients \(\Delta W\)\\
%     % \Return{\(W\)}
% \STATE \textbf{return} \(W\)

\end{algorithmic}
\end{algorithm}

% \begin{algorithm}[htb]
% \caption{EC-LDA against FGL (Client-side) \Jeff{need to compact}}
% \label{client-side}
% \textbf{Known}: local epochs \(E\), learning rate \(\eta\), local data \(G_{n}(V_{n}, \mathbb{E}_{n}, Y_{n}^{V}, Y_{n}^{\mathbb{E}})\), where \(Y_{n}^{V}\) is the node labels, \(Y_{n}^{\mathbb{E}}\) is the link labels\\
% \textbf{Input}: Global model \(W\)\\
% \textbf{Output}: Trained global model \(W\)
% % \textbf{ClientUpdata}:
% \begin{algorithmic}[1]
% % \STATE Node classification tasks\\
% \STATE Receive current global model $W$ from server
% \FOR{\(e = 1,2,\ldots,E\)}
% \STATE // Node classification tasks
% \STATE $W\leftarrow W-\eta\nabla\mathcal{L}(W,V_{n},\mathbb{E}_{n};Y_{n}^{V})$ 
% \STATE // Link prediction tasks
% \STATE $W\leftarrow W-\eta\nabla\mathcal{L}(W,V_{n},\mathbb{E}_{n};Y_{n}^{\mathbb{E}})$ 
% \ENDFOR
% \STATE \textbf{return} \(W\)
% \end{algorithmic}
% \end{algorithm}

% \section{Active Label Distribution Inference Attack against FGL}
\section{Our Methodology: EC-LDA}\label{sec:4}

% We propose the \textbf{A}ctive \textbf{L}abel \textbf{D}istribution Inference \textbf{A}ttack against FGL (ALDA), which aims to infer the label distribution information of the training data on victim clients. In this section, we will elaborate on threat model, the principles of attack, why active attack is chosen and how active attack is executed.

% We first introduce the basic LDA attack \Wendy{This is unclear. Is this basic LDA attack designed by you or is it existing in the literature?}, followed by the enhanced version, the EC-LDA algorithm.

% Based on the above analysis, we conclude that the key to reducing error lies in minimizing the variance of \(I\). To minimize the variance of \(I\), one effective approach is to constrain the absolute values of \(I_k\) within a small range, which can be achieved by clipping the model parameters.

Based on the above analysis, we conclude that the key to reducing error lies in minimizing the variance of \(I\), which can be effectively achieved by constraining the absolute values of \(I_k\) by clipping the model parameters.

\subsubsection{Overview} We now provide the introduction to EC-LDA, an enhanced attack method targeting GNNs. Unlike LDA, EC-LDA makes full use of Equation \ref{equ_10} by clipping the parameters of the global model before distributing it to clients, which reduces the variance of node embeddings, thus making the variance of \(I\) smaller. 

As shown in Fig. \ref{fig_dataflow}, EC-LDA consists of three main steps: 
\begin{itemize}
    \item \textbf{Step 1: Clipping model.} The server clips the initial model and sends it to clients to get the gradients. Algorithm \ref{clip_parameters} demonstrates the clipping method. First the \(\ell_{2}\) norm \(\mathbb{N}\) of the model is computed, and then \(p/max(1,\frac{\mathbb{N}}{C})\), i.e., the clipped parameters, will replace the original parameters \(p\). After receiving the global model sent by the server, the client trains the model using its own private training data and uploads the trained model to the server.
    % \item \textbf{Step 2: Forward propagation of dummy data.} The server generates dummy training data randomly and inputs it into the initial model to obtain \(I_{k}\), \(\bar{I}\), and \(p_{k,l}\).
    \item \textbf{Step 2: Forward propagation of dummy data.} The server first generates dummy training data, which is then fed into the initial model. Through forward propagation, the server obtains intermediate outputs \(I_{k}\), \(\bar{I}\) from the input of the final fully-connected layer. Simultaneously, the model's output is processed by the softmax layer to derive \(p_{k,l}\).
    % \item \textbf{Step 3: Calculating distribution.}  After the server gets gradients, \(p_{k,l}\), \(I_{k}\), and \(\bar{I}\), Equation \ref{equ_10} is used to infer the label distributions of client-side private data.
    
    \item \textbf{Step 3: Calculating distribution.} Following local client training and forward propagation of dummy training data, the server obtains intermediate variables gradients, \(p_{k,l}\), \(I_{k}\), and \(\bar{I}\). It then calculates the private label distribution of client-side data by applying Equation \ref{equ_10}.

\end{itemize}

\subsubsection{EC-LDA}
Algorithm \ref{active_attack_algo} presents the server-side implementation of EC-LDA within the FGL framework. The boxed part represents the additional component of EC-LDA compared to the normal training process. First, the server initializes the model parameters and attack results (Line 1). If the server does not perform an attack in current round \(r\), the training process is no different from the normal training process. Otherwise,  the server saves the initial model \(W_{r-1}\) for round \(r\) as \(W_{r-1}^{'}\) and clip \(W_{r-1}\) with Algorithm \ref{clip_parameters}. The server generates dummy training data and feeds it into \(W_{r-1}^{'}\) to obtain $p_{k,l}$, $I_k$ and $\bar{I}$ (Line 4-6). Then the server sends the clipped model down to all clients, and each client trains on its local private data and uploads the trained model. After training, server calculates the label distribution \(D\) with Equation \ref{equ_10}, and saves \(D\) into \(\mathbb{L}\) (Line 11-12). The server uses the saved initial model \(W_{r-1}^{'}\) of round \(r\) to replace \(W_{r}\) (Line 17). This prevents propagation of clipping-affected models to subsequent rounds, thereby ensuring minimal performance impact from sporadic attacks. 
The training process of clients does not differ from the general training process, clients first receive the model \(W_{r-1}\) from the server, then train the model \(W_{r-1}\) with the local privacy training data for \(E\) times and upload the trained model to the server. In other words, EC-LDA trades one round of training resources for one round of label distributions of client-side private training data.

\subsubsection{Extend to Link Prediction}
Implementing EC-LDA requires GNN layers with a fully connected final layer and the use of the cross-entropy loss function, which makes EC-LDA particularly versatile within classification tasks on GNNs, such as node classification. Since the link prediction is a binary classification task, EC-LDA is still applicable to link prediction tasks as long as the last fully connected layer of the model has two output units and the cross-entropy loss function is used. For node classification, EC-LDA can attack the label proportions, while for link prediction, it can target the graph density.

\section{Experiments}\label{sec:5}
In this section, we demonstrate the effectiveness of EC-LDA through answering the following three research questions:
\begin{itemize}
    \item RQ1 - How effective is EC-LDA under node classification and link prediction tasks with real-world graph datasets?
    \item RQ2 - How robust is EC-LDA across different experimental parameters?
    % \Wendy{"Robust" may not be accurate. Do you mean "effective" instead?}
    \item RQ3 - What is the performance of EC-LDA under differential privacy protection?
    % \item RQ3 - How well does EC-LDA perform on other classification tasks, such as link prediction tasks?
\end{itemize}
% In this section, we first introduce the experimental setups, then we conduct a series of ablation experiments to explore the effect of different factors on EC-LDA.

\subsection{Experimental Settings}

% The details of datasets and hyper-parameter settings please refer to the Appendix~\ref{Appen:dataset}. We employed distinct configurations for each dataset to prevent overfitting.

\textbf{Datasets:} We conduct experiments on six widely used public datasets: Cora \cite{sen2008collective}, CiteSeer \cite{sen2008collective}, LastFM \cite{rozemberczki2020characteristic}, Facebook \cite{rozemberczki2021multi}, CoraFull \cite{bojchevski2017deep}, and WikiCS \cite{mernyei2020wiki}. Cora, CiteSeer, WikiCS, and CoraFull are citation network datasets. Facebook and LastFM are social network datasets. We show the main features of these datasets in Table \ref{dataset_main_info}. To demonstrate the effectiveness of EC-LDA under conditions with a large number of labels, we select the CoraFull dataset, which has up to 70 labels.

\textbf{Hyper-parameter Settings:}
All datasets use the SGD optimizer. If not specified, all the experiments in this paper use the following setup: all at \(E\) of 5, clipping threshold \(C\) of 0.01, number of clients of 10, and attack on all the clients in the middle round. All the experiments in this paper are taken with different random seeds to repeat 5 times, and the results are averaged.

\begin{table}[htbp]
\caption{Statistics of datasets}
\centering
\setlength{\tabcolsep}{1mm}{
\begin{tabular}{@{}cccccc@{}}
\toprule
Types & Datasets & \#Nodes & \#Edges & \#Features & \#Classes \\ \midrule
\multirow{4}{*}{\begin{tabular}[c]{@{}c@{}}Citation\\ Network\end{tabular}} & CiteSeer & 2120    & 7358    & 3703       & 6         \\
                                                                            & Cora     & 2485    & 10138   & 1433       & 7         \\
                                                                            & WikiCS   & 11311   & 297033  & 300        & 10        \\
                                                                            & CoraFull & 18800   & 125370  & 8710       & 70        \\ \midrule
\multirow{2}{*}{\begin{tabular}[c]{@{}c@{}}Social\\ Network\end{tabular}}   & Facebook & 22470   & 342004  & 128        & 4         \\
                                                                            & LastFM   & 7624    & 55612   & 128        & 18        \\ \bottomrule
\end{tabular}}

\label{dataset_main_info}
\end{table}

\begin{table*}[t!]
\caption{Performance of EC-LDA and baselines.}
\centering
\begin{tabular}{@{}cccccccccc@{}}
\toprule
\multirow{2}{*}{Dataset}  & \multirow{2}{*}{GNN types} & \multicolumn{4}{c}{Cos-sim}                  & \multicolumn{4}{c}{JS-div}                   \\ \cmidrule(l){3-6}  \cmidrule(l){7-10}
                          &                            & EC-LDA         & Infiltrator & iLRG  & LLG*  & EC-LDA         & Infiltrator & iLRG  & LLG*  \\ \midrule
\multirow{3}{*}{CoraFull} & GCN                        & \textbf{1.000} & 0.946       & 0.299 & 0.418 & \textbf{0.000} & 0.031       & 0.400 & 0.506 \\
                          & GAT                        & \textbf{1.000} & 0.977       & 0.314 & 0.395 & \textbf{0.000} & 0.017       & 0.415 & 0.515 \\
                          & GraphSAGE                  & \textbf{1.000} & 0.946       & 0.249 & 0.304 & \textbf{0.000} & 0.035       & 0.486 & 0.552 \\  \midrule
\multirow{3}{*}{LastFM}   & GCN                        & \textbf{1.000} & 0.953       & 0.590 & 0.414 & \textbf{0.000} & 0.040       & 0.212 & 0.410 \\
                          & GAT                        & \textbf{1.000} & 0.978       & 0.494 & 0.349 & \textbf{0.000} & 0.023       & 0.281 & 0.451 \\
                          & GraphSAGE                  & \textbf{1.000} & 0.966       & 0.470 & 0.347 & \textbf{0.000} & 0.038       & 0.315 & 0.448 \\ \midrule
\multirow{3}{*}{WikiCS}   & GCN                        & \textbf{1.000} & 0.865       & 0.610 & 0.475 & \textbf{0.001} & 0.058       & 0.212 & 0.402 \\
                          & GAT                        & \textbf{1.000} & 0.987       & 0.622 & 0.488 & \textbf{0.001} & 0.010       & 0.214 & 0.396 \\
                          & GraphSAGE                  & \textbf{1.000} & 0.721       & 0.561 & 0.512 & \textbf{0.001} & 0.130       & 0.242 & 0.374 \\ \midrule
\multirow{3}{*}{Cora}     & GCN                        & \textbf{1.000} & 0.981       & 0.601 & 0.558 & \textbf{0.001} & 0.009       & 0.267 & 0.327 \\
                          & GAT                        & \textbf{1.000} & 0.995       & 0.499 & 0.527 & \textbf{0.001} & 0.004       & 0.328 & 0.343 \\
                          & GraphSAGE                  & \textbf{1.000} & 0.987       & 0.469 & 0.588 & \textbf{0.000} & 0.009       & 0.382 & 0.312 \\ \midrule
\multirow{3}{*}{CiteSeer} & GCN                        & \textbf{1.000} & 0.996       & 0.579 & 0.432 & \textbf{0.001} & 0.004       & 0.295 & 0.386 \\
                          & GAT                        & \textbf{1.000} & 0.997       & 0.522 & 0.439 & \textbf{0.001} & 0.004       & 0.294 & 0.383 \\
                          & GraphSAGE                  & \textbf{1.000} & 0.992       & 0.474 & 0.409 & \textbf{0.000} & 0.008       & 0.368 & 0.402 \\ \midrule
\multirow{3}{*}{Facebook} & GCN                        & \textbf{1.000} & 0.991       & 0.679 & 0.581 & \textbf{0.000} & 0.003       & 0.195 & 0.326 \\
                          & GAT                        & \textbf{0.999} & 0.991       & 0.670 & 0.434 & \textbf{0.002} & 0.003       & 0.208 & 0.394 \\
                          & GraphSAGE                  & \textbf{1.000} & 0.982       & 0.643 & 0.519 & \textbf{0.000} & 0.008       & 0.251 & 0.354 \\ 
\bottomrule
\end{tabular}

\label{tab:all_dataset_gnn_types_nc}
\end{table*}

\textbf{Evaluation Metrics:} Inspired by \cite{geng2016label}, we use the following two evaluation metrics to fully demonstrate the effectiveness of EC-LDA: cosine similarity (cos-sim) and Jensen-Shannon divergence (JS-div). Cos-sim measures the similarity of two distributions and is applicable when the similarity of vectors is not directly related to the length of the vectors, while JS-div measures the distance between two distributions.
The two metrics are calculated using the following two formulas:
\begin{equation}\label{equ_7}
\begin{aligned}
\text{JS-div}(D,D^{*})=\frac{1}{2}\sum_{n=1}^{N}d_n\log(\frac{2d_{n}}{d_{n}+d_{n}^{*}})\\
+\frac{1}{2}\sum_{n=1}^{N}d_{n}^{*}\log(\frac{2d_{n}^{*}}{d_{n}+d_{n}^{*}}),
\end{aligned}
\end{equation}
\begin{equation}
\begin{aligned}
\text{cos-sim}(D,D^{*})=\frac{\sum_{n=1}^{N}(d_{n}d_{n}^{*})}{\sqrt{\sum_{n=1}^{N}d_{n}^{2}}\sqrt{\sum_{n=1}^{N}d_{n}^{*2}}},
\end{aligned}
\end{equation}
where \(D\) is the inferred label distribution and \(D^*\) is the ground-truth label distribution.
Cos-sim takes values in the range \([-1,1]\). The larger the cos-sim, the closer the inferred label distribution is to the ground-truth label distribution, i.e., the more effective the attack is. The value of JS-div ranges from \([0,1]\), and the more effective the attack is, the smaller the JS-div is.

\textbf{Model Architecture:} To illustrate the general applicability of EC-LDA, we choose three classical GNN models as global models, which are GCN, GAT, and GraphSAGE. All of the above models consist of two parts, the GNN layers and the fully connected layers.

\textbf{Baselines:} We compare EC-LDA with three different attacks which are Infiltrator \cite{meng2023devil}, iLRG \cite{ma2023instance}, and LLG* \cite{wainakh2021user}. Infiltrator infers the label of the victim node by adding a neighbor to the victim node and observing the output of the neighbor. It is worth noting that Infiltrator focuses on node-level attacks, and for comparison, we attack all training nodes with Infiltrator. LLG* and iLRG both reveal the number of each label, and we use the number of samples per label extracted by iLRG and LLG* to compute the label distribution of each client. Similar to EC-LDA, iLRG and LLG* also use the gradients of the last fully connected layer of the model, and when \(E\) is greater than 1, we use \(\Delta W^{(FC)}/E\) as an approximation of the gradients.

% 通过之前的分析，我们知道I的方差越小，攻击的效果就会越好。而dummy training data也和I的方差有关，因此我们设置它由均值为0，标准差为0.001的高斯分布生成，并且包含1000个节点。这样的设置可以提升攻击的效果。
\textbf{Dummy Training Data:} Based on the previous analysis, we know that the smaller the variance of \(I\), the better the attack performance. Since the dummy training data is also related to the variance of \(I\), we generate it from a Gaussian distribution with a mean of 0 and a standard deviation of 0.001, containing 1000 nodes. 
% This configuration enhances the attack performance.

% z这里应该不用说明了
% The model used for all experiments is a 2-layer GNN connected to a fully connected layer. 

% \begin{table}[ht]
% \centering
% \begin{tabular}{@{}ccc@{}}
% \toprule
% Dataset  & Learning Rate & Global Rounds \\ \midrule
% CiteSeer & 0.1           & 100           \\
% Cora     & 0.1           & 200           \\
% WikiCS   & 0.5           & 300           \\
% CoraFull & 0.5           & 200           \\
% Facebook & 0.1           & 200           \\
% LastFM   & 0.1           & 200           \\ \bottomrule
% \end{tabular}
% \caption{Partial experimental settings for node classification tasks.}
% \label{tab:nc_seetings}
% \end{table}

% \begin{table}[ht]
% \centering
% \begin{tabular}{@{}ccccc@{}}
% \toprule
% \multirow{3}{*}{Dataset} &
%   \multicolumn{2}{c}{Node Classification} &
%   \multicolumn{2}{c}{Link Prediction} \\ \cmidrule(l){2-5} 
%  &
%   \begin{tabular}[c]{@{}c@{}}Learning\\ Rate\end{tabular} &
%   \begin{tabular}[c]{@{}c@{}}Global\\ Rounds\end{tabular} &
%   \begin{tabular}[c]{@{}c@{}}Learning\\ Rate\end{tabular} &
%   \begin{tabular}[c]{@{}c@{}}Global\\ Rounds\end{tabular} \\ \midrule
% CiteSeer & 0.1 & 100 & 0.5 & 300 \\
% Cora     & 0.1 & 200 & 1   & 200 \\
% WikiCS   & 0.5 & 300 & 0.6 & 300 \\
% CoraFull & 0.5 & 200 & 1   & 300 \\
% Facebook & 0.1 & 200 & 0.1 & 300 \\
% LastFM   & 0.1 & 200 & 0.5 & 100 \\ \bottomrule
% \end{tabular}
% \caption{Partial experimental settings for node classification task and link prediction task.}
% \label{tab:nc_lp_seetings}
% \end{table}

\subsection{Attack Performance (RQ1)}

\noindent \textbf{Node Classification.}  We evaluate the performance of EC-LDA with all datasets and all GNN types, which, for the node classiﬁcation task, have a varying number of labels and are widely distributed, with 4 kinds of labels for nodes in the Facebook dataset and 70 kinds of labels for nodes in the CoraFull dataset, the datasets with the smallest and the largest number of classes, respectively. Table \ref{tab:all_dataset_gnn_types_nc} shows the experimental results. From the comparison presented in Table \ref{tab:all_dataset_gnn_types_nc}, it is evident that EC-LDA consistently demonstrates exceptional performance across various datasets and three distinct GNN models, as indicated by the cos-sim scores consistently at or above 0.999 and the JS-div consistently at or below 0.002, aligning closely with the optimal values. Remarkably, EC-LDA’s outstanding performance remains consistent regardless of the number of labels, the number of nodes, and the specific GNN types, showcasing its broad applicability. Additionally, EC-LDA consistently outperforms other methods across all experiments.

\begin{table}[t!]
\caption{Performance of EC-LDA on link prediction tasks.}

\centering
\begin{tabular}{@{}cccc@{}}
\toprule
Dataset                   & GNN types & Cos-sim & JS-div \\ \midrule
\multirow{3}{*}{Facebook} & GCN       & 1.000   & 0.000  \\
                          & GAT       & 1.000   & 0.000  \\
                          & GraphSAGE & 1.000   & 0.000  \\ \midrule
\multirow{3}{*}{CiteSeer} & GCN       & 1.000   & 0.000  \\
                          & GAT       & 1.000   & 0.000  \\
                          & GraphSAGE & 1.000   & 0.000  \\ \midrule
\multirow{3}{*}{Cora}     & GCN       & 1.000   & 0.000  \\
                          & GAT       & 1.000   & 0.000  \\
                          & GraphSAGE & 1.000   & 0.000  \\ \midrule
\multirow{3}{*}{WikiCS}   & GCN       & 1.000   & 0.000  \\
                          & GAT       & 1.000   & 0.000  \\
                          & GraphSAGE & 1.000   & 0.000  \\ \midrule
\multirow{3}{*}{LastFM}   & GCN       & 1.000   & 0.000  \\
                          & GAT       & 1.000   & 0.000  \\
                          & GraphSAGE & 1.000   & 0.000  \\ \midrule
\multirow{3}{*}{CoraFull} & GCN       & 1.000   & 0.000  \\
                          & GAT       & 1.000   & 0.000  \\
                          & GraphSAGE & 1.000   & 0.000  \\ \bottomrule
\end{tabular}
\label{tab:all_dataset_gnn_types_lp}
\end{table}

\noindent \textbf{Link Prediction.} In the link prediction experiments, we set the same number of positive and negative edges. Table \ref{tab:all_dataset_gnn_types_lp} shows the performance of EC-LDA on all datasets in the link prediction task. In all experiments of link prediction, cos-sim and JS-div reached 1.000 and 0.000, respectively, demonstrating the stunning performance of EC-LDA.

\subsection{Ablation Experiments (RQ2)} To explore the effectiveness of EC-LDA under varying parameters, we performed extensive ablation experiments on the node classification tasks. Specifically, we explored the impact of the GNN layer depth, local epochs \(E\), clipping threshold \(C\), number of clients, dummy training data separately. 
Furthermore, we also demonstrate the label distribution of clients during training.
% Please refer to the Appendix \ref{appen:num_of_clients}  and \ref{appen:dummy_training_data} for relevant experiments regarding the number of clients and dummy training data.

\begin{figure}[ht]
    \centering
    \includegraphics[width=\linewidth]{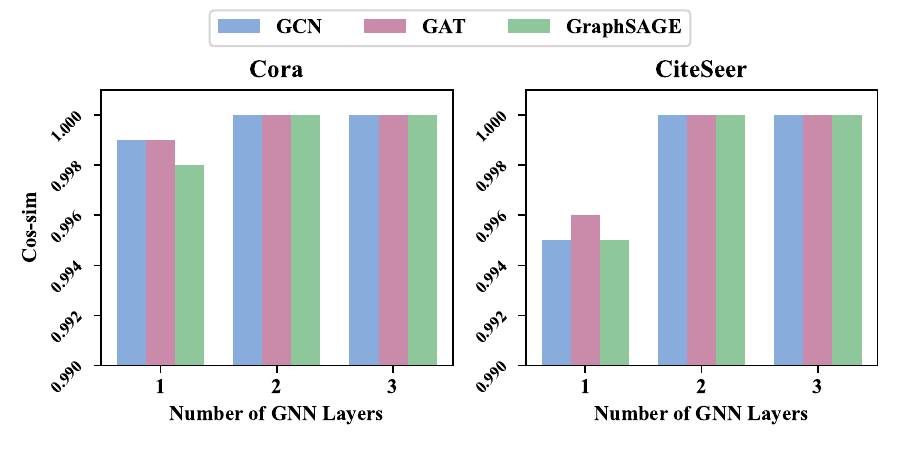}

    \caption{Performance of EC-LDA under various the number of GNN layers.}
    \label{fig:gnn_layers}
    % \vspace{5pt}
\end{figure}

\textbf{Impact of the GNN Layer Depth.} We illustrate in Fig. \ref{fig:gnn_layers} the effect of GNN layer depth on the performance of EC-LDA. Specifically, we evaluate three GNN models with 1, 2, and 3 GNN layers, respectively. Fig. \ref{fig:gnn_layers} shows that as the GNN layers increase, EC-LDA performance improves, contrary to Fig. \ref{all_dataset_var_metrics_hops}. This is due to clipping reducing model parameters, which lowers the absolute values and variance of \(I_k\), resulting in better attack performance. 
% \sout{From the figure, we can see that as the number of GNN layers increases, the attack performance of EC-LDA gets better, which can be explained by the fact that the variance of node embeddings gets smaller and smaller after the node features pass through the clipped GNN layers, leading to better and better attack performance.} 
Overall, EC-LDA delivers excellent performance across three kinds of models.

% The more GNN layers there are, the higher-order neighbor information will be included in the final node embeddings obtained. However, the number of GNN layers should not be too many because the performance of the model decreases as the number of GNN layers increases, due to the fact that the final node embeddings obtained tend to converge towards similarity.

\begin{figure}[htbp]
    \centering

    \includegraphics[width=\linewidth]{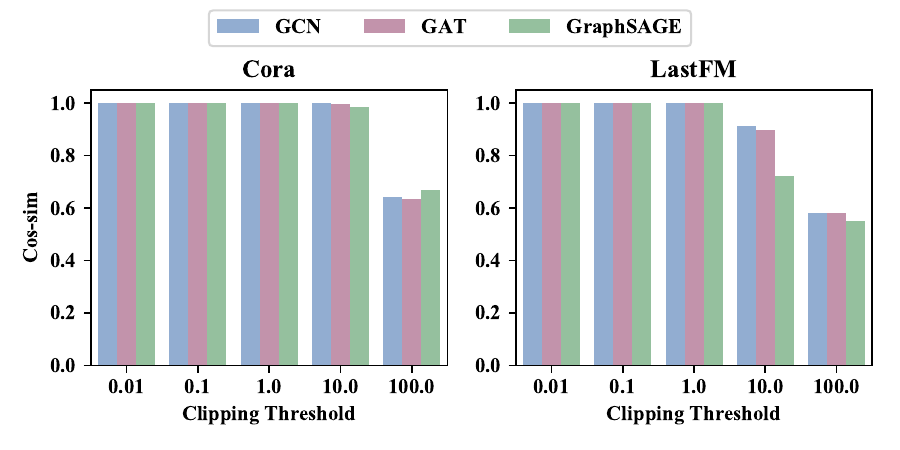}

    \caption{Performance of EC-LDA under various clipping threshold \(C\).}
    \label{fig:clipping_norm_C}
        % \vspace{-2mm}
        % \vspace{-5pt}
\end{figure}

\begin{figure*}[htbp]
    \centering

    \includegraphics[width=\linewidth]{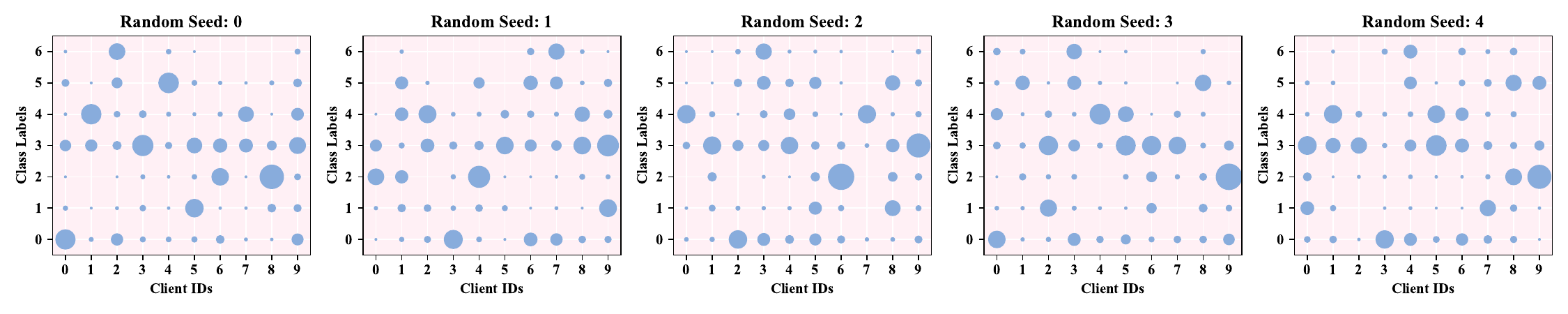}

    \caption{The label distribution obtained by each client when dividing the Cora dataset under five different random seeds.}
    \label{fig:Cora_label_dis}
\end{figure*}

\textbf{Impact of clipping threshold \(C\).} In EC-LDA, the server will distributes the clipped model during attack rounds. The method of clipping directly affects the performance of EC-LDA. We employ the \(\ell_{2}\) norm clipping method and investigate the impact of \(C\) on EC-LDA. Fig. \ref{fig:clipping_norm_C} shows that the performance of EC-LDA under different \(C\). We can observe that as \(C\) increases, the performance of EC-LDA deteriorates. This is because with increasing \(C\), the clipping intensity decreases. When \(C\) exceeds the \(\ell_{2}\) norm of the model parameters, the value of \(max(1,\frac{\mathbb{N}}{C})\) will be equal to 1, which means no clipping will be applied to the model, leading to a deterioration in EC-LDA's performance.

\textbf{Impact of local epochs \(E\).} In previous experiments we fixed \(E\) to be 5, a widely accepted practice in FGL.
To study the effect of different \(E\) on EC-LDA, we evaluate the performance of EC-LDA under different \(E\). We show the experimental results in Table \ref{tab:local_epochs_metrics}. It is evident that EC-LDA performs exceptionally well across varying \(E\), with its performance across all datasets almost remaining unaffected by changes in \(E\). As introduced in the previous section, we use \(\Delta W_{n}^{(FC)}/E\) as an approximation of the gradients for EC-LDA. As \(E\) increases, the error between the true gradients and the approximated gradients becomes larger, which results in a slight degradation of the performance of EC-LDA in some cases. Overall, EC-LDA performs effectively across different \(E\).

\begin{figure}[htbp]
    % \centering
    \centerline{
    \includegraphics[width=\linewidth]{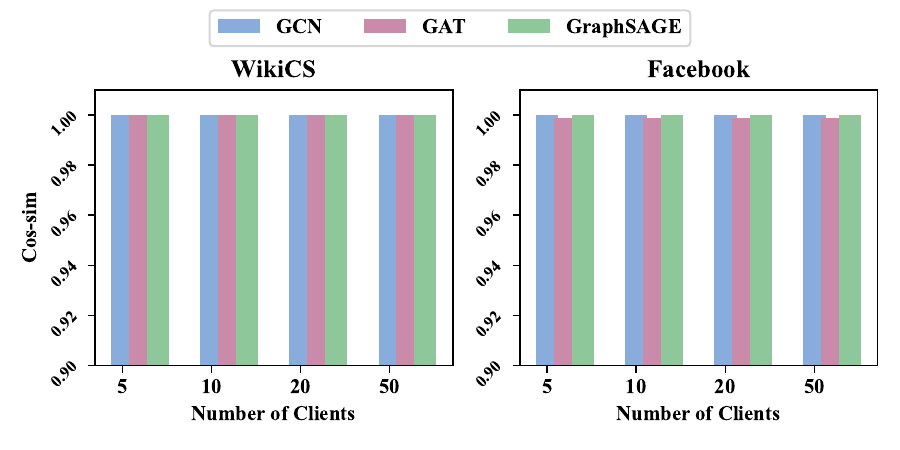}}
    \caption{Performance of EC-LDA under various client number.}
    \label{fig:num_clients}
    \vspace{-10pt}  % 下移5pt
\end{figure}

\textbf{Impact of the Client Number.} 
% \label{appen:num_of_clients}
To investigate the impact of the number of clients on the performance of EC-LDA, we conducted experiments on the Cora dataset, varying the number of clients from 5 to 50. Fig. \ref{fig:num_clients} demonstrates the performance of EC-LDA as the number of clients changes. We can see that regardless of the number of clients, the performance of EC-LDA remains excellent, showcasing its wide applicability.

% 图片n展示了EC-LDA的效果随着客户端数量的变化，我们可以看到无论客户端的数量怎么变化，EC-LDA的表现一直非常好,展示了EC-LDA的广泛适用性。

\begin{table}[t!]
\caption{Performance of EC-LDA across various local epochs \(E\).}
\centering
\setlength{\tabcolsep}{1mm}{
\begin{tabular}{@{}cccccc@{}}
\toprule
\multirow{2}{*}{Model} & \multirow{2}{*}{\begin{tabular}[c]{@{}c@{}}Local\\ Epochs\end{tabular}} & \multicolumn{2}{c}{Cos-sim} & \multicolumn{2}{c}{JS-div} \\ \cmidrule(l){3-4} \cmidrule(l){5-6} 
                           &   & Cora  & Facebook & Cora  & Facebook \\ \midrule
\multirow{3}{*}{GCN}       & 1 & 1.000 & 1.000    & 0.000 & 0.000    \\
                           & 3 & 1.000 & 1.000    & 0.000 & 0.000    \\
                           & 5 & 1.000 & 1.000    & 0.001 & 0.000    \\ \midrule
\multirow{3}{*}{GAT}       & 1 & 1.000 & 1.000    & 0.000 & 0.000    \\
                           & 3 & 1.000 & 1.000    & 0.000 & 0.001    \\
                           & 5 & 1.000 & 0.999    & 0.001 & 0.002    \\ \midrule
\multirow{3}{*}{GraphSAGE} & 1 & 1.000 & 1.000    & 0.000 & 0.000    \\
                           & 3 & 1.000 & 1.000    & 0.000 & 0.000    \\
                           & 5 & 1.000 & 1.000    & 0.000 & 0.000    \\ \bottomrule
\end{tabular}}

\label{tab:local_epochs_metrics}
\end{table}

\textbf{Impact of Dummy Training Data.}
% \label{appen:dummy_training_data}
In EC-LDA, dummy training data is employed to generate parameter \(K\), \(p_{k,l}\), \(I_{k}\), and \(\bar{I}\).
Table \ref{tab:dummy_training_data_standard_dev} illustrates the impact of the standard deviation in dummy training data on the performance of EC-LDA. The results indicate that changes in the distribution of dummy training data have minimal influence on the effectiveness of EC-LDA.

\begin{table}[t!]
% \vspace{-5pt}  % 上移5pt
\caption{Performance of EC-LDA under various standard deviation of dummy training data. }
\centering
\footnotesize
\begin{tabular}{cccc}
\toprule
Dataset                   & Standard Deviation & Cos-sim & JS-div \\ \midrule
\multirow{3}{*}{CiteSeer} & 0.001              & 1.000   & 0.001  \\
                          & 0.1                & 1.000   & 0.001  \\
                          & 10                 & 1.000   & 0.001  \\ \midrule
\multirow{3}{*}{Facebook} & 0.001              & 1.000   & 0.000  \\
                          & 0.1                & 1.000   & 0.000  \\
                          & 10                 & 1.000   & 0.000  \\ \bottomrule
\end{tabular}

\label{tab:dummy_training_data_standard_dev}
% \vspace{-5pt}  % 下移5pt
\end{table}

\textbf{Impact of Label Distribution.}
We employ a community detection algorithm to partition the graph dataset for each client, which identifies communities within the network(i.e., groups of nodes with high connection densities). Fig. \ref{fig:Cora_label_dis} shows the distribution of labels assigned to each client by this algorithm for the Cora dataset. From the figure, it is evident that, with a setup of 10 clients, the algorithm simulates a variety of label distributions. Our prior experimental results indicate that EC-LDA consistently delivers strong performance. Therefore, we conclude that EC-LDA can perform well across a range of distributions.

\begin{figure*}[htbp]
    \centering

    \includegraphics[width=\linewidth]{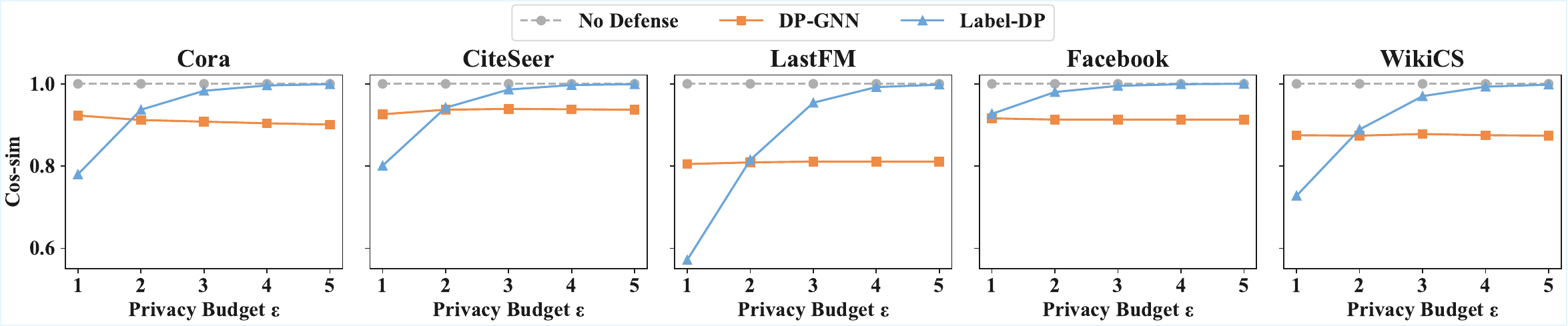}
    % \caption{The attack performance in EC-LDA, DP-GNN and Label-DP.}
    \caption{The attack performance of EC-LDA under no defense, DP-GNN defense, and Label-DP defense.}
    \label{fig_nc_DP}
    % \vspace{-5mm}
\end{figure*}

\begin{figure*}[htbp]
    \centering

    \includegraphics[width=\linewidth]{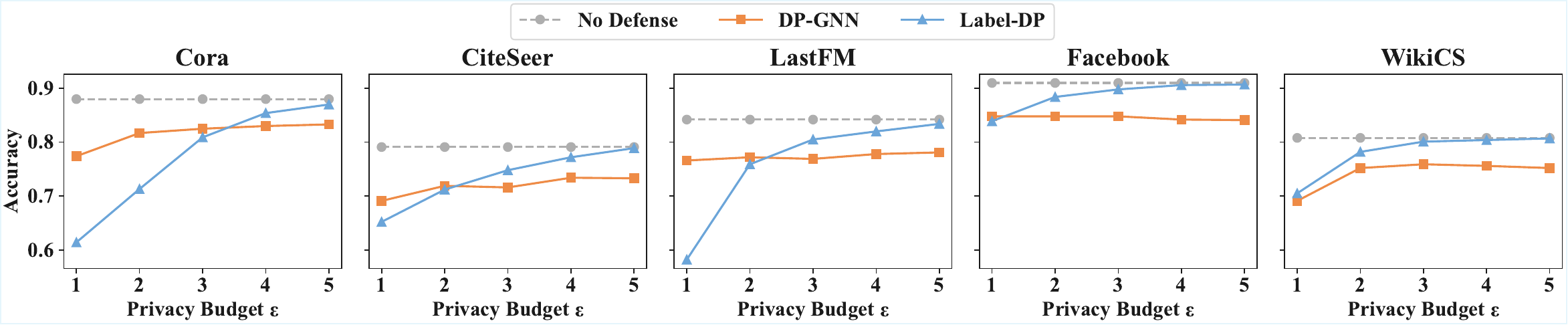}
    % \caption{The model accuracy in EC-LDA, DP-GNN and Label-DP.}
    \caption{The model accuracy when executing EC-LDA attacks under no defense, DP-GNN defense, and Label-DP defense.}
    \label{fig_nc_DP_acc}
        % \vspace{-5mm}
\end{figure*}

\subsection{Defense Performance of Differential Privacy (RQ3)}
Differential privacy (DP) is a powerful privacy-preserving technology widely used in machine learning due to its rigorous mathematical definition. DP can defend against various attacks, such as membership inference attacks \cite{hui2021practical}, adversarial example attacks \cite{lecuyer2019certified}, and data reconstruction attacks \cite{balle2022reconstructing}. Specifically, we consider node-level differential privacy DP-GNN \cite{daigavane2021node} and label-level differential privacy Label-DP \cite{ghazi2021deep}. DP-GNN controls the out-degree of nodes and is a variant of DP-SGD \cite{abadi2016deep,fu2024dpsur}. Label-DP adds noise to the label matrix to protect the labels of the training dataset.
In DP, privacy budget $\epsilon$ controls the scale of the added noise: a smaller privacy budget $\epsilon$ results in larger noise, and vice versa. 
Fig. \ref{fig_nc_DP} and Fig. \ref{fig_nc_DP_acc} respectively present the performance of EC-LDA and the node classification accuracy of the FGL model under DP-GNN and Label-DP defenses.

% This improvement arises because Label-DP introduces increasing noise to the labels as privacy budget $\epsilon$ decreases. 
\textbf{Label-DP.}
From Fig. \ref{fig_nc_DP}, it is evident that the defense effectiveness of Label-DP improves with decreasing privacy budget $\epsilon$. 
Under the Label-DP defense, noise is injected into the label matrix of the training dataset. As the $\epsilon$ value decreases, the magnitude of noise added to the label matrix progressively increases. This results in a widening divergence between the perturbed label matrix and the ground-truth label matrix.
Additionally, the inferred label distribution closely matches the distribution after noise addition due to EC-LDA's powerfulness in recovering label distributions. 
% Critically, while EC-LDA accurately reconstructs the \textit{noise-corrupted} label distribution, this reconstructed distribution intrinsically inherits the perturbations introduced during defense.
% Critically, EC-LDA's reconstructed distribution inherits the defensive perturbations.
This alignment results in a significant disparity between the inferred distributions and actual label distributions, consequently diminishing EC-LDA's attack effectiveness. 
% distributions, consequently diminishing EC-LDA's attack effectiveness. 
% distributions, consequently diminishing EC-LDA's attack effectiveness. 

\textbf{DP-GNN.} 
% However, different with Label-DP defense, as privacy budget $\epsilon$ decreases, the defense effect of DP-GNN remains almost unchanged, 
However, as shown in Fig. \ref{fig_nc_DP}, unlike Label-DP, the defensive efficacy of DP-GNN remains nearly unchanged as \(\epsilon\) decreases from 5 to 1.
Contrary to the well-established observation that diminishing $\epsilon$ enhances privacy protection, DP-GNN exhibits anomalous $\epsilon$-invariance. 
This unexpected phenomenon stems directly from the gradient processing mechanics of DP-GNN.
As a variant of DP-SGD, DP-GNN first clips the gradients and then injects zero-mean Gaussian noise $\mathcal N(0,\sigma^2)$ into gradients.
However, when EC-LDA calculates the label distribution, the gradient it uses is \(\Delta W_{l}^{(FC)}\), which is the sum of the gradients of the \(l\)-th output unit in the last fully connected layer along the input dimension, and 
\(\Delta W_{l}^{(FC)}=\sum_{m}\Delta W_{m,l}^{(FC)}\).
We define \(\Delta \overline{W}_{l}^{(FC)}\) as the clipped version of gradient \(\Delta W_{l}^{(FC)}\), and \(\Delta W_{l}^{'(FC)}\) as the gradient after adding noise to \(\Delta \overline{W}_{l}^{(FC)}\). We have:

\begin{equation}
\begin{aligned}
% \nabla z_{k,n}\mathcal{L}(v_{k},y_{k}) &=\frac{e^{z_{k,n}}}{\sum_{n^{\prime}=1}^{N}e^{z_{k,n^{\prime}}}}-y_{k,n} \\
% &=p_{k,n}-y_{k,n}
\Delta W_{l}^{'(FC)}
&=\sum_{m}(\Delta \overline{W}_{m,l}^{(FC)}+\eta_{m,l}),
% &=\sum_{m}\Delta \overline{W}_{m,l}^{(FC)}+\sum_{m}\eta_{m,l},
% &=\Delta W_{l}^{(FC)}+\eta_{l},
\end{aligned}
\end{equation}
where \(\eta_{m,l} \sim \mathcal N(0,\sigma^2)\), then \( \mathbb{E}[\eta_{m,l}]=0\), in that way:
\begin{equation}
\begin{aligned}
&\mathbb{E}[\Delta W_{l}^{'(FC)}-\Delta \overline{W}_{l}^{(FC)}]\\
&=\mathbb{E}[\sum_{m}\Delta \overline{W}_{m,l}^{(FC)}]+\mathbb{E}[\sum_{m}\eta_{m,l}]-\mathbb{E}[\sum_{m}\Delta \overline{W}_{m,l}^{(FC)}]=0.
\end{aligned}
\end{equation}
It can be interpreted as: during the summation of \(\Delta W_{l}^{'(FC)}\) over the input dimensions, the Gaussian noise added by DP-GNN undergoes positive-negative cancellation, causing the summed gradient after noise addition, i.e., \(\Delta W_{l}^{'(FC)}\), to approximate the clipped gradient, i.e., \(\Delta \overline{W}_{l}^{(FC)}\). Consequently, this leads to the defensive efficacy of DP-GNN being insensitive to \(\epsilon\).

\section{Related Work}\label{sec:6}
\textbf{Inference Attack against GNNs.} \cite{chen2023empirical} investigated graph data leakage in horizontal federated and vertical federated scenarios. Specifically, link inference attack and attribute inference attack are proposed in the vertical federated scenario, graph reconstruction attack and graph feature attack are proposed in the horizontal federated scenario. \cite{qiu2022your} proposed an inference attack against the relationships between nodes of GNNs in vertical federated scenarios. \cite{meng2023devil} inferred the labels of victim nodes by infiltrating the raw graph data, i.e., by adding elaborately designed nodes and edges, and their attacks can only be performed on trained models. However, their attacks focus on the node-level, and there are no extant papers on label distribution inference attacks at the graph-level for FGL. 

\textbf{Label Recovery Attack.} The label distribution inference attack is a type of label recovery attack. \cite{zhu2019deep} was the first to restore the training sample from gradients. They restored the input data and associated label from the gradients using gradient-matching. This method continuously optimizes the dummy input data and associated label by minimizing the mean square error of the gradients of the dummy sample with respect to the true gradients.
\cite{zhao2020idlg} introduced iDLG and were the first to propose that with a non-negative activation function, privacy label can be extracted with 100\% success rate from the signs of gradients at the output layer. Both of these methods are applicable only to single-sample training scenarios.
Afterwards, \cite{yin2021see,geng2021towards} extended the attacks to the mini-batch scenario. 
% 有很多工作都是从梯度出发来获取隐私标签信息。
Extensive research such as \cite{ma2023instance,dang2021revealing,zhou2022ppa,wainakh2021user,gu2023ldia} demonstrates that gradients can be exploited to extract private label distributions.
% Among them,
% \cite{ma2023instance} recovered class-wise embeddings from the gradients and further restored the number of each label.
% \cite{dang2021revealing} proposes RLG, which extracts label information with the gradients of the output layer, but RLG requires the use of the soft-max activation function. Additionally, RLG only reveals which labels are used for training, without disclosing the quantity of samples per label. 
% \cite{zhou2022ppa} proposed PPA, which can infer a client's label information but can only output the majority class or minority class. It cannot fully reflect the label information.
% \cite{wainakh2021user} proposed LLG, which utilizes the magnitude and direction of shared gradients to determine whether a specific label is present. LLG has three versions, suitable for different scenarios.
% Additionally,
\cite{aggarwal2021label} relies on number theory and combinatorics to recover label information from log-loss scores. However, their method is sensitive to the number of classes in the dataset, and as the variety of labels increases, not only does the accuracy decrease, but the inference time also increases.
% \cite{ma2023instance} recovered class-wise embeddings from the gradients and further restored the number of each label.
\cite{wang2019eavesdrop} proposed three attack methods, which can infer whether a specific label appears in the training process, the quantity of each label for a specific client in a round, and the quantity of each label throughout the entire training process, respectively.
% One method can infer whether a specific label appears in the training process, while the other two methods can infer the exact quantity of each label. One method can infer the quantity of each label for a specific client in a round, while the other method can infer the quantity of each label throughout the entire training process.

Overall, most existing label recovery attacks focus on image datasets, and none of them discuss label distribution on graph datasets. In this paper, We focus on label distribution inference attack against GNNs.
% which can more fully represent the privacy information of the training dataset.

\section{Limitations} \label{sec:7}
In EC-LDA, the malicious server enhances attack effectiveness by clipping to compress the variance of node embeddings, then distributing this clipped model parameters to clients. This clipping operation leads to an anomalous distributional compression of model parameters. Consequently, a potentially effective defense strategy is for clients to detect potential attacks by identifying these abnormally compressed global model parameters and implementing corresponding countermeasures. However, since EC-LDA only requires a single clipping operation and can be executed at any point during the FGL process, this presents significant challenges for client-side detection and defense. For future work, we will explore effective detection and defense methods against EC-LDA. Additionally, we will concentrate on enhancing the stealthiness of our attacks. 
To enhance stealth, the server can deploy a ``fishing model" that mimics the original model's performance and parameter distribution while achieving the same effect as embedding compression, allowing the attack to proceed covertly while maintaining its effectiveness.
% This could involve training a "fishing model" that mimics the performance and parameter distributions of the original model for client deployment, thereby replacing the compressed model. 

\section{Conclusion}\label{sec:8}
% This paper demonstrates the effectiveness of EC-LDA for label privacy in FGL scenarios.

This paper presents EC-LDA, a label distribution inference attack designed for Federated Graph Learning (FGL). The proposed method was developed by investigating the impact of node embedding variance in Graph Neural Networks (GNNs) on existing LDA approaches. By compressing node embeddings, EC-LDA significantly enhances the attack capability of a malicious server. Extensive experiments on six graph datasets show that EC-LDA outperforms state-of-the-art methods in both node classification and link prediction tasks. Additionally, we demonstrate the robustness of EC-LDA under various parameter settings and under DP protection.

\section{ACKNOWLEDGMENTS}
This work is supported by Fundamental Research Funds for the Central Universities (Grant No. 40500-20104-222609), National Natural Science Foundation of China Key Program (Grant No. 62132005), Natural Science Foundation of Shanghai (Grant No.22ZR1419100), and CAAI-Huawei MindSpore Open Fund (Grant No.CAAIXSJLJJ-2022-005A).

%% The file named.bst is a bibliography style file for BibTeX 0.99c
\bibliographystyle{IEEEtran}
\bibliography{ijcai25}

\clearpage

% \appendix
\appendices

\end{document}